%% file: main.tex
\documentclass[10pt,twocolumn,letterpaper]{article}

\usepackage[pagenumbers]{cvpr} % To force page numbers, e.g. for an arXiv version

\input{preamble}

\definecolor{cvprblue}{rgb}{0.21,0.49,0.74}
\usepackage[pagebackref,breaklinks,colorlinks,citecolor=cvprblue]{hyperref}
\usepackage{cuted}
\usepackage{capt-of}
\usepackage{bm}

\def\paperID{308} % *** Enter the Paper ID here
\def\confName{3DV\xspace}
\def\confYear{2025\xspace}

\title{Geometry-guided Cross-view Diffusion for One-to-many Cross-view Image Synthesis}

\author{%
Tao Jun Lin$^{1}$, Wenqing Wang$^{2}$, Yujiao Shi$^{3}$, \\Akhil Perincherry$^{4}$, Ankit Vora$^{4}$ and Hongdong Li$^{1}$\\
$^1$The Australian National University \quad $^2$University of Surrey \\
$^3$ShanghaiTech University \quad $^4$Ford Motor Company\\
\texttt{taojun.lin@anu.edu.au, 
shiyj2@shanghaitech.edu.cn}\\
}
\vspace{-5mm}

\begin{document}
\maketitle
\input{sec/0_abstract}

\input{sec/1_intro}
\input{sec/2_related}
\input{sec/3_methods}

\input{sec/4_experiments}
\input{sec/5_discussion}

{
    \small
    \bibliographystyle{ieeenat_fullname}

}
\input{sec/X_suppl}

\end{document}

%% file: preamble.tex
\usepackage[dvipsnames]{xcolor}

%% file: sec/0_abstract.tex
\begin{strip}
\centering
   \includegraphics[width=0.85\textwidth]{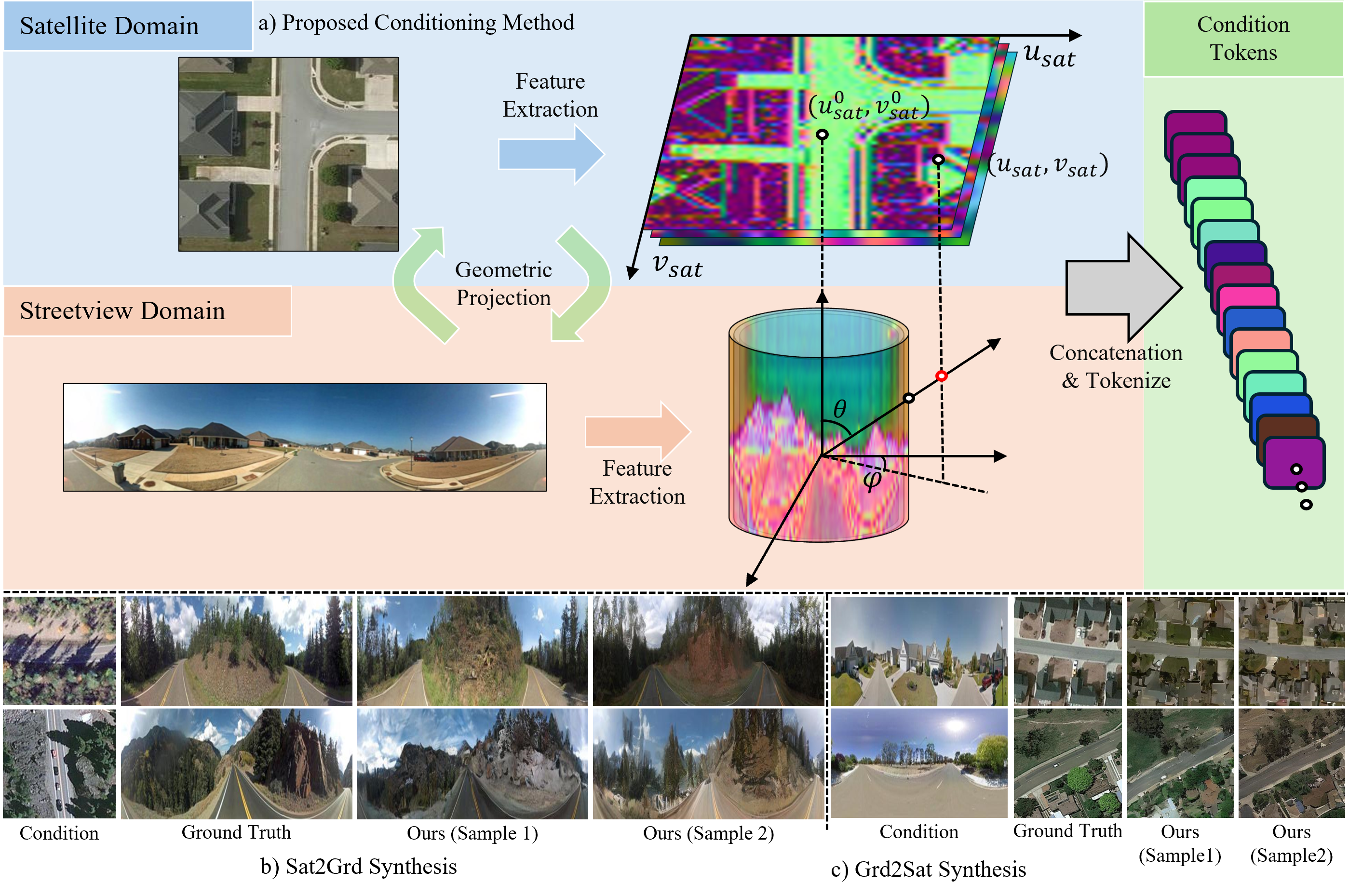}
\captionof{figure}{Our proposed Geometry-Guided Conditioning method for cross-view image
synthesis (a) and visualization examples generated by our proposed Geometry-guided Cross-view Diffusion. On the bottom left (b) are images generated from our \textbf{Sat2Grd} model, on the bottom right (c) are image generated from our \textbf{Grd2Sat} model.
\label{fig:teaser}}
\vspace{-3mm}
\end{strip}

\begin{abstract}
This paper presents a novel approach for cross-view synthesis aimed at generating plausible ground-level images from corresponding satellite imagery or vice versa. We refer to these tasks as satellite-to-ground (Sat2Grd) and ground-to-satellite (Grd2Sat) synthesis, respectively. Unlike previous works that typically focus on one-to-one generation, producing a single output image from a single input image, our approach acknowledges the inherent one-to-many nature of the problem. This recognition stems from the challenges posed by differences in illumination, weather conditions, and occlusions between the two views. To effectively model this uncertainty, we leverage recent advancements in diffusion models. Specifically, we exploit random Gaussian noise to represent the diverse possibilities learnt from the target view data. We introduce a Geometry-guided Cross-view Condition (GCC) strategy to establish explicit geometric correspondences between satellite and street-view features. This enables us to resolve the geometry ambiguity introduced by camera pose between image pairs, boosting the performance of cross-view image synthesis. Through extensive quantitative and qualitative analyses on three benchmark cross-view datasets, we demonstrate the superiority of our proposed geometry-guided cross-view condition over baseline methods, including recent state-of-the-art approaches in cross-view image synthesis. Our method generates images of higher quality, fidelity, and diversity than other state-of-the-art approaches. 
\end{abstract}

%% file: sec/1_intro.tex
\section{Introduction}

\label{sec:intro}
Ground-and-satellite cross-view image synthesis has attracted considerable attention recently due to its potential applications in virtual reality, simulations, cross-view image matching and data augmentation, \etc. The task is to synthesize a target view image from a given viewpoint and a relative pose between the two views. The synthesized images are expected to not only exhibit a geometrically consistent scene structure between the views but also maintain high visual fidelity to real-world data. 

The cross-view image synthesis is a remarkably challenging and inherently ill-posed learning task. This complexity arises primarily from the drastic viewpoint change, which results in minimal Field-of-View (FoV) overlap, severe occlusion, and large discrepancies in image contents and visual features. Preliminary works in cross-view synthesis mostly relied on conditional Generative Adversarial Networks \cite{mirza2014conditional}. Some of them focus on generating corresponding ground-view conditioned on a given satellite image patch, employing high-level semantics or contextual information for supervision \cite{regmi2018cross, regmi2019cross,tang2019multi, zhai2017predicting, lu2020geometry}. 
Recent research \cite{li2021sat2vid, shi2022geometry, Sat2Density} has further proven that incorporating 3D geometry into the learning process can significantly boost the quality of generated ground-view images. 
However, all these works formulate the task as a deterministic image-to-image translation, while the ground-and-satellite cross-view synthesis is inherently a probabilistic one-to-many problem.  

Diffusion models have emerged as a powerful new family of deep generative models and have achieved state-of-the-art results in generative tasks, especially in image generation \cite{song2020score, dhariwal2021diffusion, ho2020denoising}. The recent Latent Diffusion models (LDM) \cite{dhariwal2021diffusion} have enabled the probabilistic generation of high-quality images from any prompts, making it a preferable option to model the uncertainty in the ground-and-satellite cross-view synthesis task.

Most of the recent researches follow the path of Text-to-Image generation, utilizing superior power of vision-language models such as CLIP~\cite{radford2021learning}. Zero123 \cite{liu2023zero1to3} demonstrates a way to prepare an image condition with its camera pose information by concatenating image CLIP encoding and frequency embedded camera pose. Then using it as a conditioning representation for fine-tuning the pre-trained Stable Diffusion Model to learn posed CLIP embeddings. This conditioning method has demonstrated promising performance in the multi-view synthesis task at object level, but the model needs to implicitly learn the relationship between the conditioning image, pose, and the target image. We have discovered that image CLIP embedding is insufficient in generating cross-view images with fine-grained geometric accuracy and spatially alignment due to underlying ambiguous relationship between the ground-view image and camera pose, please see \cref{sec:cond} for more details.

To address the ambiguity mentioned above, we propose Cross-view Diffusion, a conditioned cross-view synthesis framework developed upon LDM, as described in Fig. \ref{fig:overview}. Instead of using the widely accepted pretrained CLIP image encoder\cite{ye2023ipadaptertextcompatibleimage, liu2023zero1to3}, this paper leverages a Geometry-guided Cross-view condition that is derived from explicit 3D geometric projection of extracted image features. The proposed condition demonstrates capability in generating cross-view images of high quality and fidelity with fine-grained geometric and semantic control. Importantly, with the same framework, our method is able to handle both ground-to-satellite and satellite-to-ground image synthesis, where Grd2Sat is considered more challenging \cite{regmi2018cross} due to the limited FoV of ground images and the presence of occlusion in ground views. 
The main contributions of this paper are summarized as follows:
\begin{itemize}
    \item We present Geometry-guided Cross-view Diffusion, a conditional generative framework for cross-view image synthesis. Our approach showcases state-of-the-art performance in synthesizing images for both Sat2Grd and Grd2Sat tasks across multiple cross-view datasets. 
    \item We propose a Geometry-guided Cross-view Conditioning approach, a novel 3D geometry-aware condition to guide the generation process of diffusion models. This conditioning approach eases the burden of diffusion models without implicitly learning the cross-view domain discrepancy and pose ambiguity from ground cameras, allowing our framework to generate geometrically and semantically consistent cross-view images. 
    \item Our method is able to generate plausible target-view images with diversity from single input view, successfully modeling the one-to-many property of the ground-and-satellite cross-view image synthesis task.  

\end{itemize}

%% file: sec/2_related.tex
\section{Related Work}
\begin{figure*}[h]
   \centering
   \includegraphics[width=\linewidth]{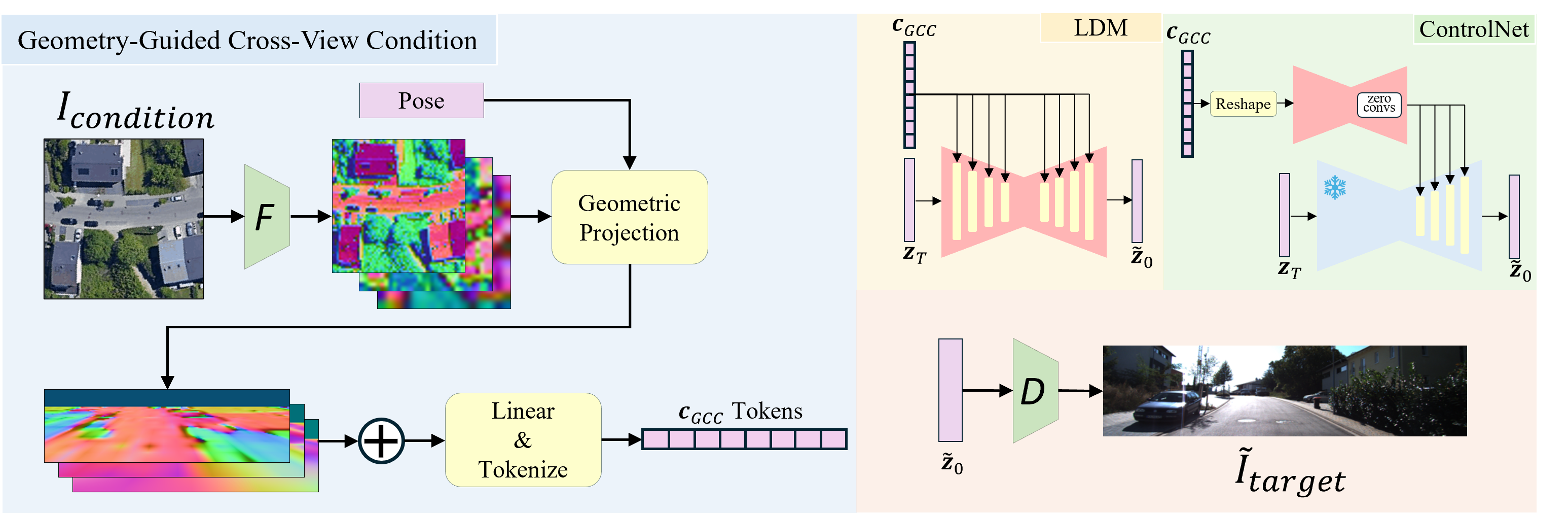}
   \caption{An overview of the proposed Cross-view Image Synthesis Pipeline. When provided with either a satellite image patch or a street-view image, the model employs a feature extractor $\mathcal{F}$ and our Geometry projection Module to construct our Geometry-guided Cross-view Conditions(GCC). The Latent Diffusion Pipeline learns to model cross-view data distribution from a Gaussian noise latent, under the guidance of our proposed GCC. The ControlNet module takes GCC as input and fine-tunes LoRA layers.}
   \label{fig:overview}
   \vspace{-5mm}
\end{figure*}

\subsection{Ground-and-satellite image synthesis }
Mapping ground and satellite images from one domain to another was first explored by Zhai~\etal~\cite{zhai2017predicting}. They proposed to learn a linear transformation matrix between satellite and street-view semantics. Later, Regmi and Borji~\cite{regmi2018cross, regmi2019cross} demonstrated that the conditional GANs could effectively address the ground-and-satellite cross-view image synthesis task, and adding an additional branch to the network for semantic map estimation could facilitate the view synthesis quality.
After that, different powerful networks have been exploited for this task~\cite{tang2019multi,wu2022cross,zhu2023cross}.

Recently, researchers explored how to combine the geometric correspondences between the views for the cross-view synthesis task~\cite{lu2020geometry,li2021sat2vid}. Lu~\etal~\cite{lu2020geometry} proposed first estimating height and semantic maps from satellite images, which was then used to recover the ground structure and prepare a better condition for the ground-view image synthesis. 
Considering this method requires GT height and semantic maps of satellite images for training, Shi~\etal~\cite{shi2022geometry} modeled geometric correspondences between the cross-view pixels in an end-to-end framework, eliminating the need for semantic and satellite map height supervision.  The most recent work, Sat2Density \cite{Sat2Density}, further improved the synthesis quality by modeling the transparency and illumination of sky regions. 

However, all the above works formulate the task as a deterministic mapping. The ground-and-satellite cross-view synthesis is inherently a one-to-many task due to the severe occlusions and illumination differences between the views. This paper resorts to the recently advanced diffusion models to address the probabilistic nature. 

\subsection{Cross-view image-based localization}
Ground-to-satellite camera localization aims to determine a ground camera's location against a satellite map. The task was proposed initially for city-scale localization and formulated by cross-view image retrieval~\cite{castaldo2015semantic, lin2013cross, mousavian2016semantic, workman2015location, workman2015wide, vo2016localizing, Cai_2019_ICCV, yang2021cross, Zhu_2022_CVPR, Hu_2018_CVPR, Liu_2019_CVPR, sun2019geocapsnet, zhu2021revisiting, Regmi_2019_ICCV, shi2019spatial, shi2020optimal, toker2021coming, shi2020looking, vyas2022gama, shi2023cvlnet, zhang2023cross}. 
In this task, many works~\cite{Regmi_2019_ICCV,toker2021coming} have demonstrated that the cross-view image synthesis objective is beneficial to improve the cross-view localization performance. 
Recently, researchers have extended the task to fine-grained pose refinement once the most similar satellite image has been retrieved for the query image~\cite{shi2022beyond, xia2022visual, xia2023convolutional, lentsch2023slicematch, fervers2023uncertainty, sarlin2023orienternet, shi2023boosting, zhang2024increasing, xia2025adapting, shi2025weakly}. 
In this line of work, the cross-view feature synthesis has been extensively explored. Shi and Li~\cite{shi2022beyond} exploited satellite-to-ground feature synthesis because ground images have a larger resolution of scene contents and thus is sensitive to camera location change.  Fervers~\etal~\cite{fervers2023uncertainty}, Shi~\etal~\cite{shi2023boosting} and OrienterNet~\cite{sarlin2023orienternet} leveraged ground-to-satellite synthesis as registering synthesized overhead view feature map to reference satellite feature map leads to more efficient camera pose computation.  
Instead of cross-view feature synthesis, this paper addresses the problem of cross-view image synthesis.
We expect this task can potentially facilitate cross-view localization performance by introducing more readily available cross-view image pairs for localization network training. 

%% file: sec/3_methods.tex
\section{Methods}
\subsection{Problem Formulation} 
This paper focuses on synthesizing corresponding cross-view images conditioned on either a ground-view image or a satellite image, so we address the task as a conditional image generation problem. Given a satellite image patch $I_s \in \mathbb{R}^{H_s \times W_s \times 3}$ or a ground image $\mathbb{R}^{H_g \times W_g \times 3}$ along with a relative pose between the cross-view input-target images, we expect our model to learn to synthesize corresponding ground image $I'_g \in \mathbb{R}^{H_g \times W_g \times 3}$ or satellite image $I'_s \in \mathbb{R}^{H_s \times W_s \times 3}$, respectively, under the guidance of the conditioned input. The dimensions of the image prompt and desired generation target from the same domain are kept identical for simplicity.

We then follow LDM \cite{rombach2022highresolution} to train our model in a learned image latent space, a perceptually equivalent space to the data space, but is more suitable for likelihood-generative models and more computationally efficient. We aim to reconstruct the latent $\mathbf{z}_{0}$ from a Gaussian noise sample $\mathbf{z}_{T} \sim \mathcal{N}(\mathbf{0},\mathbf{I})$, where $\mathbf{z}_{0} = \mathcal{E}(\mathbf{x}_{0})$ is a desired data point encoded by a learnt image encoder. Subsequently, the training objective of the denoising process is formulated as:
\begin{equation}
    \mathcal{L}_{LDM} := \mathbb{E}_{\mathbf{z}\sim\mathcal{E}(\mathbf{x}),\mathbf{c}_{GCC},{\epsilon}\sim\mathcal{N}(\mathbf{0},\mathbf{I}), t}
    \bigg[
    \|{\epsilon}-{\epsilon}_\theta(\mathbf{z}_t,t,\mathbf{c}_{GCC})\|^2_2\bigg],
\end{equation}
which is guided by our proposed Geometry-guided Cross-view Condition $\mathbf{c}_{GCC}$. Overview of the framework is shown in \cref{fig:overview}, more details on LDM and its formulations will be included in the supplementary material.

\subsection{Geometry-guided Cross-view Condition}

In the context of image synthesis, the relationship between the generative power of diffusion models and the representation of conditions is still an under-explored area of research \cite{rombach2022highresolution}. In this paper, we address cross-view image synthesis as a conditional image generation task. We aim to explore image synthesis conditioned on multiple guidance from different domains: a correspondent cross-view image pair with the ground camera's relative pose to the satellite image.  

To tackle the ambiguity caused by the relative camera pose between the two views, we propose Geometry-guided Cross-view Condition (GCC), a novel approach that effectively embeds camera pose information into our multi-level image features using our Cross-View Geometry Projection (CVGP) Module denoted as $\mathcal{P}$. 
The CVGP module projects the given viewpoint image feature representations to the target viewpoint according to the relative camera pose. The projected multi-level features are adopted as the diffusion model condition. 
This approach bridges the domain gap between the conditioning image and pose prompt and the desired distribution for target images.

\subsubsection{Cross-View Geometry Projection} 
\label{sec:proj}
In this section, the geometric projection from satellite image to ground panoramic image is used as an example to illustrate the pixel mapping process. We begin by defining the geometric relationship involving the reference world coordinate system, ground camera pose, and the center of the reference satellite image. In this coordinate system, its origin is set to the reference satellite image center, the $x$ axis aligns with latitude or the $v_s$ direction, the $y$ axis points downward into the image,  the $z$ axis is parallel to longitude or the $u_s$ direction. Then, we can map any point $[x, y, z]^T$ in the world coordinate system to a satellite pixel coordinate with an orthogonal projection:
\begin{equation} \label{eq:ortho}
    [u_s, v_s]^T = [\frac{z}{\gamma} + u^0_s, \frac{x}{\gamma} + v^0_s]^T,
\end{equation}
where $\gamma$ is the per-pixel real-world distance of the satellite feature map, and $[u^0_s, v^0_s]$ is the center of the satellite feature map.
Given the location correspondence between the street-view camera position and the satellite image center, we are able to derive the pixel correspondences with cross-view geometry. We define a cylindrical image plane to represent an omnidirectional street-view image, with its pixels parameterized under a spherical coordinate system. For any satellite pixel $p_s = [u_s, v_s]^T$ that is visible from the ground view, we derive the pixel mapping to its projected pixel $p_g=[\theta, \phi]^T$ in the ground-view panorama with height $z$:

\begin{align}
\begin{split}
    \theta &= \begin{cases}
    \text{atan2}(\sqrt{(v_s-v^0_s)^2+(u_s - u^0_s)^2}, \frac{-y}{\gamma})& \text{if $y \neq 0$}\\
    \pi/2 & y = 0
  \end{cases}
   \\ \phi &= \begin{cases}
    \text{atan2}(v_s-v^0_s, u_s - u^0_s) & \text{if $u_s \neq u^0_s$}\\
    \pi/2 \cdot \text{sign}(v_s-v^0_s) & u_s = u^0_s
  \end{cases}
\end{split}
\end{align}

Likewise, we can derive the mapping from a ground pixel to a satellite pixel and accommodate other camera projection models, such as a pinhole camera. Please refer to \cref{sec:add_geo} for extensive details.

\subsubsection{Multi-Level Projected Feature Aggregation}
We utilize an arbitrary feature extraction backbone to extract visual features at multiple levels for preserving both high-level semantic information and of the input image at both coarse and fine levels. The geometric projection derived in the previous section only approximates the pixel correspondence, which might be prone to distortions and misalignment. Therefore, we project the extracted multi-level deep features rather than RGB pixels for robustness towards defects caused by our geometric assumptions, such as the ground camera height and ground plane Homography. Our extensive experiments also prove that using projected RGB images as the condition results in limited performance.

In the Grd2Sat image synthesis pipeline, we begin by extracting ground feature representation $\mathbf{F}_g \in \mathbb{R}^{H_g^l \times W_g^l \times C^l}$, where $l = 1,\dots,L$ denotes the level of features. We then project these feature representations by our geometry projection module. Subsequently, we interpolate the feature maps at each level to a unified uv grid with bilinear interpolation, resulting in projected features $\mathbf{F}_{g2s}^l \in \mathbb{R}^{H_c \times W_c \times C^l}$. The final Geometry-guided Cross-view Condition is obtained by linearly mapping the aggregation of projected features from every level to the condition dimension:
\begin{equation}
    \mathbf{c}_{g2s} = Linear(\mathcal{P}({F}_g^1) \oplus \dots \oplus \mathcal{P}({F}_g^L)) \in \mathbb{R}^{H_c \times W_c \times C_c}.
\end{equation}
where the total number of visual condition token equals $H_c \times W_c$, each token with condition dimension $C_c$. 

The condition preparation procedure for the Sat2Grd image synthesis pipeline stays analogous, differing only in the height and width of the extracted features. The condition is then processed into tokens, containing multi-resolution contextual and textural information. Utilizing the cross attention mechanism in diffusion models, these tokens can be regarded as visual sentences, enforcing visual and spatial consistency in the generated imagery content.

%% file: sec/4_experiments.tex
\section{Experiments and Results}

\subsection{Dataset} 
We evaluate the performance of our method and compare it with other SOTA cross-view image synthesis methods on several benchmark datasets, including the cross-view KITTI\cite{geiger2013vision}, CVUSA \cite{zhai2017predicting} and aligned CVACT \cite{shi2022geometry} datasets. 

The cross-view KITTI dataset is splitted into three subsets, one training set and two test sets. We train our model with Training set and to ablate the performance of various conditions on both Training and Test1 sets, which are collected from the same region. We use the left ground image and its corresponding satellite image as a cross-view image pair.

CVUSA and CVACT are both cross-view datasets with panoramic ground images. CVUSA contains 35,532 location aligned cross-view image pairs for training and 8,884 image pairs for testing. However, the street-view image in CVUSA dataset are cropped at the top and bottom by Zhai \etal \cite{zhai2017predicting}, with unspecified portions.  Aligned CVACT is the location aligned split proposed for cross-view image synthesis task, it contains 26,519 training pairs and 6,288 testing pairs, which is processed to remove the misalignment between cross-view image pairs. During training and testing, we approximate the street-view image in the CVUSA dataset as having a 90$^\circ$ vertical field of view (FoV) with the central horizontal line corresponding to the horizon, and consider CVACT ground images with 180$^\circ$ horizontal visualization.

\subsection{Implementation Details}
In our implementation, we train our Sat2Grd model with 256 $\times$ 256 resized satellite image (approximately 50x50 m$^2$ ground coverage) and Grd2Sat model with 128 $\times$ 512 street-view image as our condition view. The synthesized cross-view images are 128 $\times$ 512 street-view images and 256 $\times$ 256 satellite patches for a fair comparison with existing approaches \cite{regmi2018cross, shi2022geometry, Sat2Density}. We follow Shi et al \cite{shi2022geometry} to approximate the height of the street-view camera as 1.65 meters for the KITTI dataset, and 2 meters for the CVUSA and CVACT datasets when we perform our geometric projections. 

To clarify, the LDM model in our results refers to a diffusion model trained from scratch with our proposed GCC, and the ControlNet \cite{zhang2023adding} model refers to a module finetuned based on pretrained Stable Diffusion 2.1 model\cite{rombach2022highresolution} with proposed GCC. We implement our model based on Latent Diffusion Model's \cite{dhariwal2021diffusion} architecture with feature dimension of 768, and our image latents are obtained with a pretrained VAE image encoder \cite{kingma2014auto}. LDM models are trained with a batchsize of 48 on two NVIDIA GeForce RTX 3090 for 500 epoches for LDM and 200 epoches for ControlNet. We set $T = 50$ as DDIM \cite{song2020denoising} sampling steps when inferencing samples. For extracting proposed GCC, we adopt Swin Transformer V2 \cite{liu2022swin} as our feature extractor and we construct our geometry-guided cross-view guidance with feature output from block 1, 3 and 5 from Swinv2. Furthermore, we utilize pinhole camera projection model for KITTI data and spherical camera model for CVUSA and CVACT in our Cross-view Geometry Projection Module. 
\subsection{Evaluation Metrics} In this research, We adopt two pixel-wise similarity and two learned feature similarity as metrics for quantitative evaluation. The structure similarity index measure (SSIM) and peak signal-to-noise ratio (PSNR) for measuring the pixel-wise similarity between two images. We further use Learned Perceptual Image Patch Similarity (LPIPS) \cite{zhang2018unreasonable} to evaluate the feature similarity of generated and real images. In our ablation study, we adopt VGG backbone, and for fair comparison with other methods, we employ the pretrained AlexNet \cite{krizhevsky2012imagenet} and Squeeze \cite{iandola2016squeezenet} networks as backbones for the LPIPS evaluation, denoted as $P_{\text{alex}}$ and $P_{\text{squeeze}}$, respectively. We also include FID \cite{heusel2018gans} as a measure for the similarity between our generated images and the real images from our datasets, which is proven to be consistent with increasing disturbances and human judgment.

\subsection{Ablation Study}
\label{sec:ablation}

\subsubsection{Effectiveness of Geometry-Guided Cross-view Condition}
In this section, we first conduct detailed experiments on all three datasets to validate the importance of geometry guidance in diffusion model based cross-view image synthesis. Experiments were implemented to test for following conditions: \textit{Original Image, Projected Image, Projected Feature, and proposed Geometry-guided Cross-view Condition(GCC).} The Original Image condition is obtained by encoding the original image with CLIP \cite{radford2021learning}, the Projected Image condition refers to geometric projected original RGB image, and the Projected Feature is obtained by projecting the last hidden state of the feature encoder. 

\begin{table}[h]
  \setlength{\abovecaptionskip}{0pt}
  \setlength{\belowcaptionskip}{0pt}
  \caption{Ablation study on significance of Geometric Guidance with the KITTI Dataset, under both Camera-aligned and North-aligned Setting.}
  \label{table:cond_ablat_kitti}
  \centering
  \scalebox{0.55}{
  \begin{tabular}{ c c c c c c c c c }
    \toprule
                 &                   &            &       KITTI train   &   &    & KITTI test1 & \\
          Condition Prompt &  PSNR$\uparrow$ & SSIM$\uparrow$ & LPIPS$\downarrow$  &PSNR$\uparrow$ & SSIM$\uparrow$ & LPIPS$\downarrow$\\
    \midrule
            &   Original Image  &   17.4063	& 0.3194  &  0.1968  &   \textbf{16.6424}  &  \textbf{0.2915}  &   \textbf{0.2226}\\
    Camera-Aligned&   Projected Image  &   15.0128 & 0.2030 &  0.3399  &  14.2573  &  0.1805  &  0.3903\\
            &   Projected feature  & \textbf{17.9851} & \textbf{0.3511} & \textbf{0.1824}  &   16.4536  &  0.2851 &  0.2558\\

    \midrule
            &   Original Image  &   14.533 &	0.1677  &  0.3434  &  13.893 &  0.152  & 0.3894 \\
        North-Aligned    &   Projected Feature  & \textbf{17.082} & \textbf{0.2860} & \textbf{0.1896} &   \textbf{15.274} & \textbf{0.2131} & \textbf{0.3106}\\
    \bottomrule
\end{tabular}
}
\vspace{-4mm}
\end{table}

\begin{table}[t]
  \setlength{\abovecaptionskip}{0pt}
  \setlength{\belowcaptionskip}{0pt}
  \caption{Comparison between our proposed GCC and other baseline conditions with {Sat2Grd} synthesis on CVUSA and CVACT datasets.}
  \label{table:cond_ablat_cvusa}
  \centering
  \scalebox{0.6}{
  \begin{tabular}{ c c c c c c c c }
    \toprule
            &  \multicolumn{3}{c}{CVUSA}   &   \phantom{adc}  & \multicolumn{3}{c}{CVACT} \cr
                 \cmidrule{2-4} \cmidrule{6-8} 
          Condition Prompt &  PSNR$\uparrow$ & SSIM$\uparrow$ & LPIPS$\downarrow$ & & PSNR$\uparrow$ & SSIM$\uparrow$ & LPIPS$\downarrow$ \\
    \midrule
               Original Image  &  12.078 & 0.3229  &  0.5488  & & 13.46 & 0.3815 & 0.4728 \\
               Projected Feature  &   12.550 & 0.3363 &  0.4858 & & 13.118 & 0.3355 & 0.4536 \\
               GCC (LDM)  & 14.032 & \textbf{0.3589} & \textbf{0.4665}  & &  15.286 & 0.4332 & 0.4592 \\
               GCC (ControlNet) & \textbf{14.274} & 0.3420 & 0.5063 & & \textbf{15.609}&  \textbf{0.4485} & \textbf{0.4136} \\
    \bottomrule
\end{tabular}
}
\vspace{-4mm}
\end{table}

\begin{figure*}[t]
   \centering
     \setlength{\abovecaptionskip}{0pt}
  \setlength{\belowcaptionskip}{0pt}
   \includegraphics[width=0.8\linewidth]{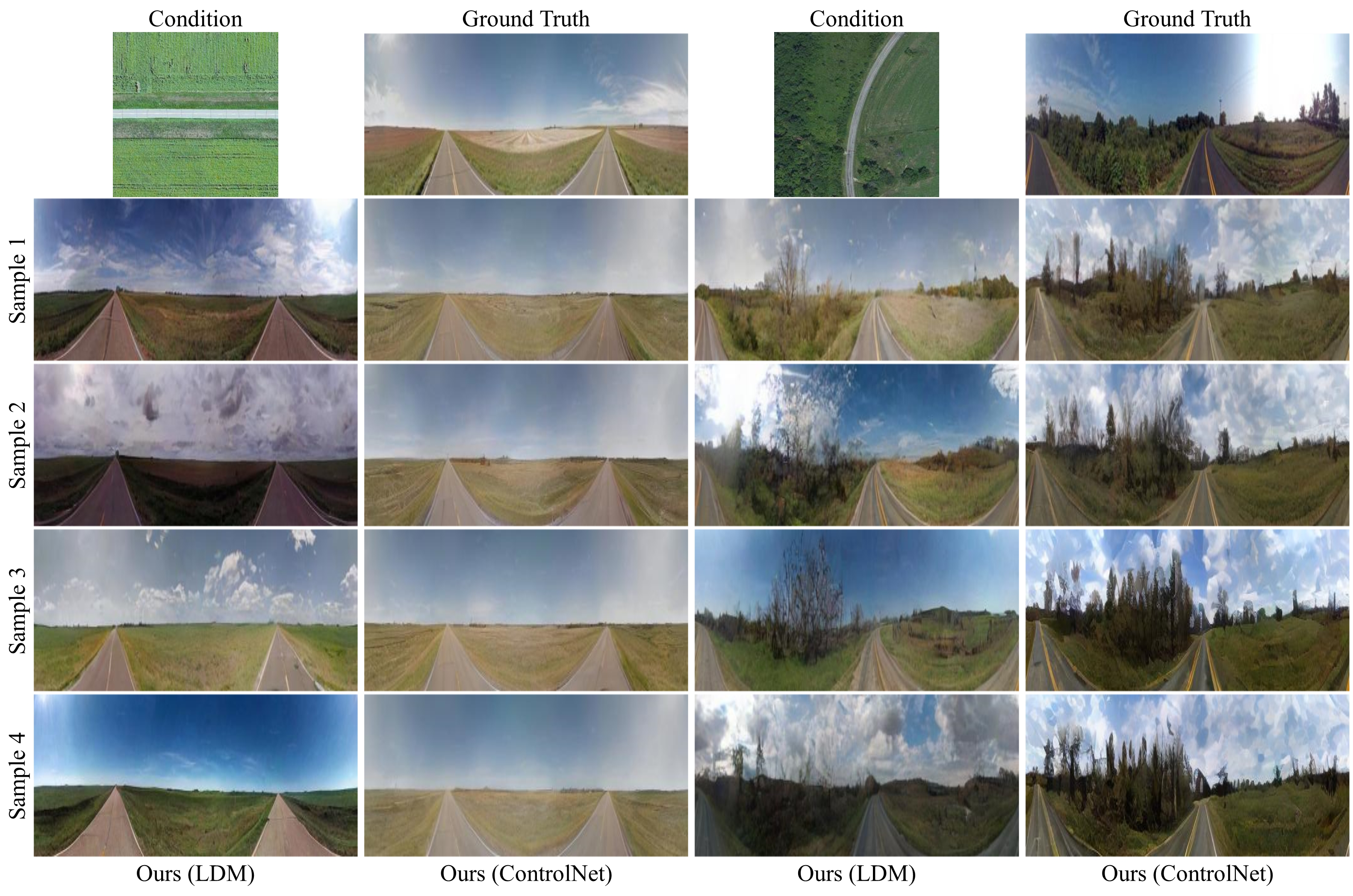}
   \caption{Ablation for sample generated by our LDM model and our ControlNet model given the same condition, on \textbf{Sat2Grd} task. }
   \label{fig:ldmvcn}
   \vspace{-4mm}
\end{figure*}
\label{sec:cond}

\begin{figure*}[h]
   \centering
     \setlength{\abovecaptionskip}{0pt}
  \setlength{\belowcaptionskip}{0pt}
   \begin{subfigure}[t]{0.85\linewidth}
       \includegraphics[width=1.0\linewidth]{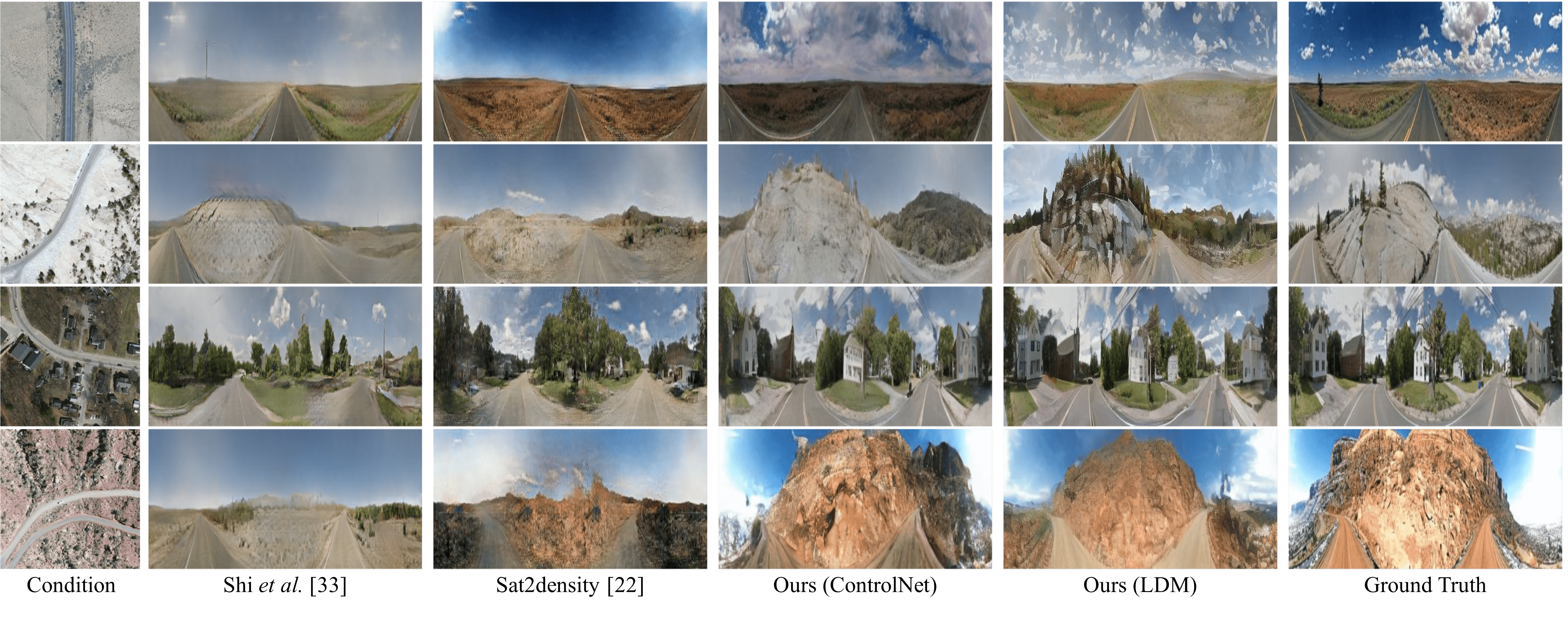}
       \caption{CVUSA Sat2Grd Visualization}
       \vspace{0.2cm}
       \label{fig:sat2grd_cvusa}
   \end{subfigure}
   
   \begin{subfigure}[t]{0.85\linewidth}
       \includegraphics[width=1.0\linewidth]{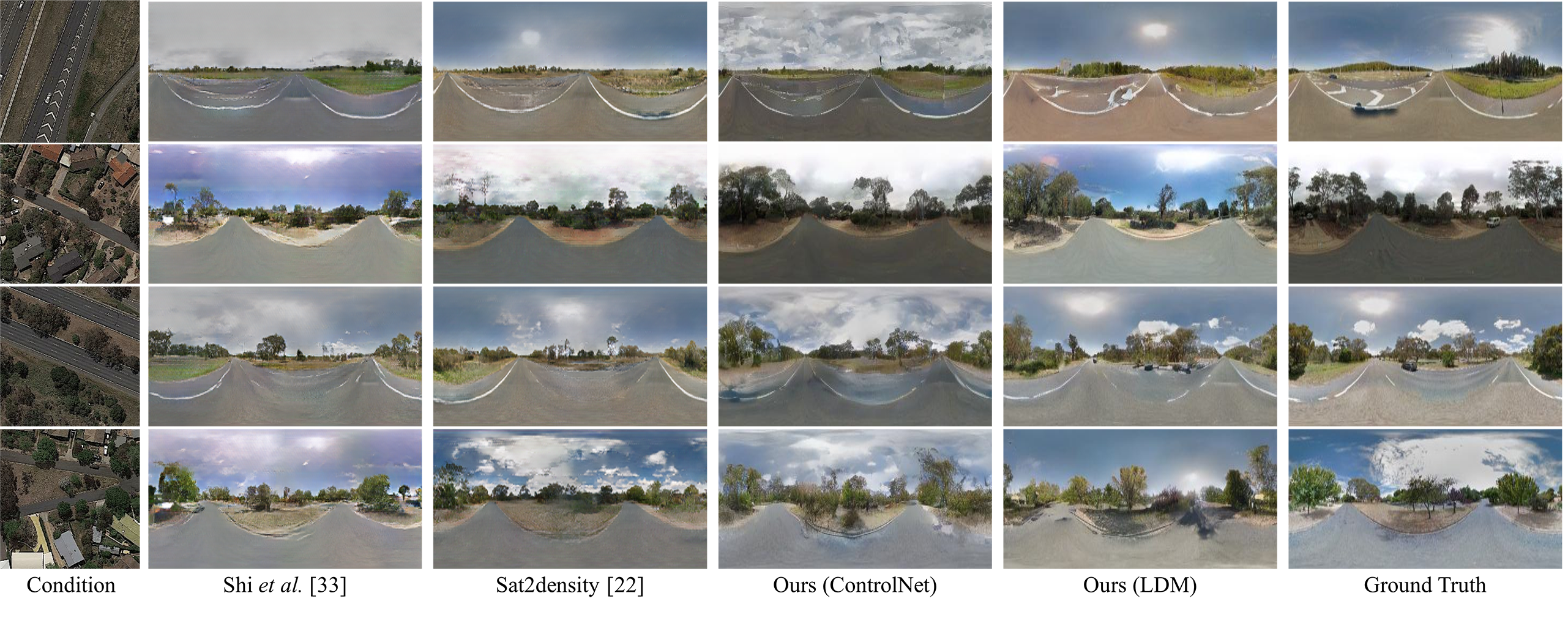}
       \caption{CVACT Sat2Grd Visualization}
       \label{fig:sat2grd_cvact}
   \end{subfigure}

   \caption{Example of generated images by different methods in \textbf{Sat2Grd} image synthesis task, on the CVACT (Aligned) and CVUSA datasets. }
   \label{fig:sat2grd}
   \vspace{-2mm}
\end{figure*}

\begin{figure}[h]
  \centering
    \setlength{\abovecaptionskip}{0pt}
  \setlength{\belowcaptionskip}{0pt}
    \includegraphics[width=1.0\linewidth]{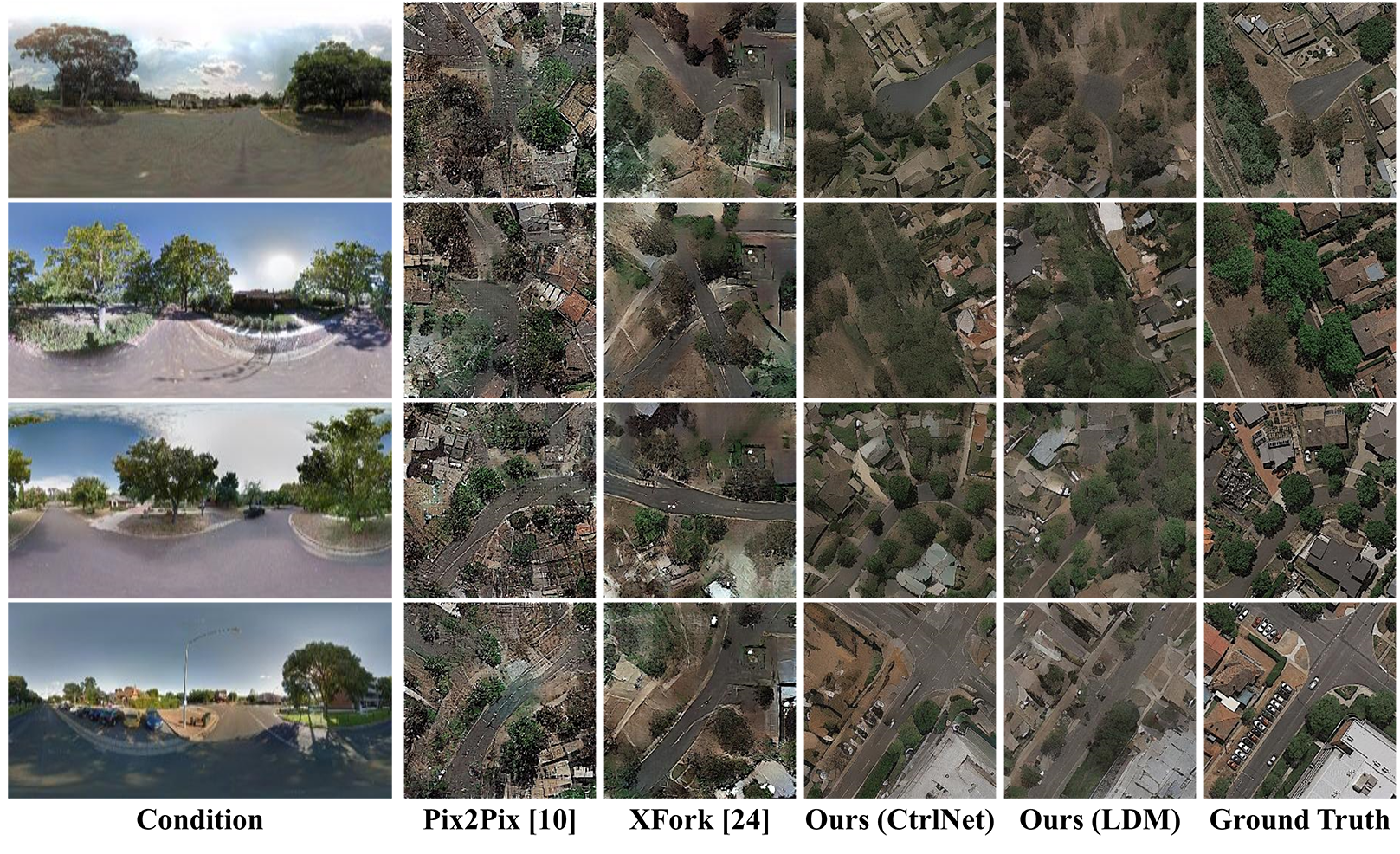}
  \caption{Example of generated images by different methods in \textbf{Grd2Sat} image synthesis
task, on the CVACT (Aligned) dataset.}
  \label{fig:grd2sat}
  \vspace{-3mm}
\end{figure}

\noindent \textbf{Camera Alignment and Pose Ambiguity} We present quantitative comparisons in \cref{table:cond_ablat_kitti} on KITTI dataset to justify the importance of geometry information in our condition design. This analysis is conducted within the \textbf{Grd2Sat} synthesis setup, where we evaluate the model's performance in generating Camera-aligned and North-aligned satellite images with various conditions. please see \cref{sec:align} for details about Camera-aligned and North-aligned setting.

As shown in the upper section of \cref{table:cond_ablat_kitti}, when the satellite image is Camera-aligned, using original image as condition can achieve comparable performance as the projected feature. However, when the satellite image is north aligned, the geometric relationship between the ground camera and the satellite remains unresolved. This pose ambiguity might hinder the model to establish connection between the condition and the data distribution. The North-Aligned results exhibit a significant decline in performance when using the original image as a condition, whereas the projected feature demonstrates robustness against the introduction of pose ambiguity. Hence, it can be suggested that providing a geometrically aligned condition is able to enhance the model's ability to learn to associate the condition and the learned data distribution.

\noindent \textbf{Capability of Conditions} 
To further assess the effectiveness of conditions, we conduct a comparative analysis of their performance by training LDM and ControlNet models on the \textbf{Sat2Grd} task using CVUSA and CVACT datasets. From \cref{table:cond_ablat_cvusa}, the utilization of our proposed condition significantly improves all three metrics for both LDM and ControlNet models. Meanwhile, LDM is still capable of learning the hidden spatial information from its structure and generate images with consistent semantic content compared to the ground truth image. This validates our assumption that tokenized projected feature sequence is capable of building dense spatial correspondence with the input latent with the cross-attention mechanism, even without any positional hints. \vspace{-2mm}
\begin{table*}[t]
  \setlength{\abovecaptionskip}{0pt}
  \setlength{\belowcaptionskip}{0pt}
  \caption{Quantitative comparison with existing  \textbf{Sat2Grd} image synthesis algorithms on the CVACT and CVUSA datasets.}
  \label{table:s2g_quantitative}
  \centering
  \scalebox{0.8}{
  \begin{tabular}{c c c c c c c c c c c c}
    \toprule
        &  & \multicolumn{3}{c}{CVUSA}   &  & \phantom{adc}  &  & \multicolumn{3}{c}{CVACT}   & \\
                 \cmidrule{2-6} \cmidrule{8-12}
       Method   &  PSNR$\uparrow$ & SSIM$\uparrow$ & $P_{\text{alex}}\downarrow$ & $P_{\text{squeeze}}\downarrow$ & $FID\downarrow$& & PSNR$\uparrow$ & SSIM$\uparrow$ &    $P_{\text{alex}}\downarrow$ & $P_{\text{squeeze}}\downarrow$ & $FID\downarrow$ \\
           \midrule
       Pix2Pix \cite{isola2017image}	& 	13.48  &  0.2946  &  0.5092 & 0.3902 & -  & &  14.38 & 0.3852 & 0.4654 & 0.3096 & - \\
       XFork \cite{regmi2018cross}	&	13.68 & 0.2873  & 0.5144 & 0.4041 & - & & 14.50 & 0.3710 & 0.4638 & 0.3262 & - \\
        Shi \etal \cite{shi2022geometry} &  13.77 &  0.3451  &  0.4639 & 0.3506 &  44.092 & & 14.59 & 0.4272 & 0.4059 & 0.2708 & 49.401\\
        Sat2Density \cite{Sat2Density} &  13.78 &  0.3301 &  0.4504 & \textbf{0.3365} &  38.078 & & 14.92 & \textbf{0.4586} & 0.3842 & 0.2573 &  38.029\\

           \midrule

        \textbf{Ours (LDM)}  &  14.032 & \textbf{0.3589} &  0.4410 & 0.3414 & 17.501 &  &  15.286 & 0.4332  &   0.3915 &  0.2669 & \textbf{21.638} \\
        \textbf{Ours (CtrlNet)} & \textbf{14.274} & 0.3420 &  \textbf{0.4345} & 0.3397 &  \textbf{13.755} & & \textbf{15.609} & 0.4485 & \textbf{0.3765} & \textbf{0.2550} & 23.706 \\

    \bottomrule
\end{tabular}
}
\vspace{-4mm}
\end{table*}

\begin{table*} [t]
  \setlength{\abovecaptionskip}{0pt}
  \setlength{\belowcaptionskip}{0pt}
  \caption{Quantitative comparison with existing \textbf{Grd2Sat} image synthesis algorithms on the CVACT and CVUSA datasets.}
  \label{table:g2s_quantitative}
  \centering
  \scalebox{0.8}{
  \begin{tabular}{ c c c c c c c c c c c c}
    \toprule
                 &         &        \multicolumn{3}{c}{CVUSA}   &  & \phantom{adc}  &  & \multicolumn{3}{c}{CVACT}   & \\
                 \cmidrule{2-6} \cmidrule{8-12}
       Method   &  PSNR$\uparrow$ & SSIM$\uparrow$ & $P_{\text{alex}}\downarrow$ & $P_{\text{squeeze}}\downarrow$ & $FID\downarrow$& & PSNR$\uparrow$ & SSIM$\uparrow$ &    $P_{\text{alex}}\downarrow$ & $P_{\text{squeeze}}\downarrow$ & $FID\downarrow$ \\        
       \midrule
      Pix2Pix \cite{isola2017image}& 11.33  &  0.1229  &  0.5490 & 0.3931 & 162.505  & & 11.60 & 0.0462 & 0.6692 & 0.3462 & 70.168\\
       % XSeq   &                       \\
       XFork\cite{regmi2018cross}&	10.85 & 0.1037 &  0.5908 & 0.4301 & 156.252 &  & 11.63 & 0.0656 & 0.6811 & 0.3716 & 184.283\\
       % Image Conditioned LDM &	11.51  &  0.1684	&  0.5446 & & & & 13.76& 0.09735 & 0.5854 \\
        \midrule
        \textbf{Ours (LDM)}   & 12.838  & 0.1751 & 0.5093 & 0.4045 & 70.125 & & 14.537 & \textbf{0.1384} & 0.4998 & 0.2561 & 53.646 \\
        \textbf{Ours (CtrlNet)} & \textbf{14.070} &  \textbf{0.2271} & \textbf{0.4829} & \textbf{0.2866} &   \textbf{53.080} & & \textbf{14.598} & 0.1375 & \textbf{0.4849} &  \textbf{0.2455} & \textbf{33.560} \\
    \bottomrule
\end{tabular}
}
\vspace{-2mm}
\end{table*}
\subsubsection{Generative Diversity of LDM and ControlNet}
The main reason we present results from both LDM and ControlNet is that the two models demonstrate different level of diversity in the \textbf{Sat2Grd} task, as shown in \cref{fig:ldmvcn}. For fair comparison, we disable classifier-free guidance scale for both models when inference. The LDM model trained from scratch is able to offer reasonable variations to the generated image, in terms of illumination, scene object and sky. While the ControlNet model generates images with less diversity but finer-grained similarity to the target image. Additionally, we noticed that both models seem to demonstrate limited generative diversity on the \textbf{Grd2Sat} task. We attribute this phenomenon to the property of data: image contents captured by satellite images are not as diverse as the ground view images, especially the sky itself can offer variety of illumination. Furthermore, geometric projection from ground-view to satellite view is ill-posed, ground features can only provide limited information that aligns with the satellite image, hence it is a harder task to learn compared to \textbf{Sat2Grd}. Please see \cref{sec:diff} and \cref{sec:fur_abla} for further explanation and examples for the two models.
\vspace{-2mm}

\subsection{Comparison with existing methods}

In the \textbf{Sat2Grd} view synthesis task, we compare our methods with Pix2Pix \cite{isola2017image}, XFork \cite{regmi2018cross}, Shi~\etal\cite{shi2022geometry} and Sat2Density \cite{Sat2Density}. As for \textbf{Grd2Sat} view synthesis, we compare our methods with Pix2Pix \cite{isola2017image} and XFork \cite{regmi2018cross}. 
Pix2Pix and XFork are classic conditional GAN-based models designed for learning image-to-image translation. However, they do not incorporate the relationship between cross-view images with 3D geometry. Shi et al and Sat2Density are both geometry-guided synthesis model. The former represents 3D geometry using a depth probability multiplane image, while the latter introduces a framework to learn a density field representation from cross-view image pairs and synthesis ground-view panoramas based on learned 3D cross-view geometry. A notable advantage of our method is that we do not require additional input such as segmentation maps and accurate height information to construct our condition. \vspace{-2mm}

\subsubsection{Quantitative Comparison} 
We report the average metrics of \textbf{10} generated samples per image condition. As presented in \cref{table:s2g_quantitative}, it is evident that our models outperform other methods on low-level and perceptual similarity measures such as SSIM, $P_{\text{alex}}$, $P_{\text{squeeze}}$ and FID in the \textbf{Sat2Grd} task. The relatively low PSNR is also in line with our expectations due to the non-deterministic nature of diffusion models. Instead, we anticipate our model to demonstrate more diversity in the generated samples, while maintaining promising structural integrity and high level alignment. It is worth noting that the Sat2Density-oracle model generates ground-view images with sky histogram from the ground truth image. Given that the sky region typically occupies nearly half of the ground-view images, possessing the ground truth sky histogram and illumination hints grants it a significant advantage in terms of pixel similarity. Therefore, simply comparing our results against theirs solely based on PSNR would be unfair. In the supplementary material, we also provide evaluation results without sky region against Sat2Density.

The quantitative results of \textbf{Grd2Sat} are displayed in \cref{table:g2s_quantitative}. Our models also achieve the best performance on all metrics, almost doubled the SSIM score of the existing researches. In both tasks, we show a notable improvement on FID score. This proves that our model is capable of learning the probabilistic distribution of the scene objects in the datasets, instead of learning one-to-one relationship to generate image only matches its ground truth target. As a result, we can generate image with high fidelity and close to real satellite and street view images.
\vspace{-3mm}

\subsubsection{Qualitative Comparison}
\cref{fig:sat2grd} displays the \textbf{Sat2Grd} synthesis results, Shi~\etal\cite{shi2022geometry} and Sat2Density \cite{Sat2Density} can generate reliable 3D geometry like road direction, while our method can clearly display road lines and better predict invisible side facade and obstacle. The last three rows in \cref{fig:sat2grd_cvusa} shows that our method can successfully synthesize building and landscape in the scene, while other two methods failed. In CVACT samples displayed in \cref{fig:sat2grd_cvact}, we show that our geometry and spatial alignment is more accurate.

\cref{fig:grd2sat} displays generated images on \textbf{Grd2Sat}, we find that Pix2Pix \cite{isola2017image}, XFork \cite{regmi2018cross} is incapable of synthesizing reasonable satellite images' structure, but our method generates consistent geometry based on the ground view, like the road intersections, shape and direction of the roads. In addition, our model is able to provide reasonable prediction for the unseen region, to generate style-consistent building and plants along the street road. \vspace{-1mm}

%% file: sec/5_discussion.tex
\section{Conclusion}

This paper has presented a novel approach to cross-view image synthesis, addressing the inherent challenges of one-to-many mapping and uncertainty modeling. By leveraging the probabilistic diffusion models and establishing explicit geometric correspondences between views, we have demonstrated significant improvements in view synthesis quality in both ground-to-satellite and satellite-to-ground synthesis across various datasets. Moving forward, extending our methodology to incorporate additional modalities, such as text, depth information or learning across multiple datasets, could broaden its applicability and enhance its capabilities.

%% file: sec/X_suppl.tex
\clearpage
\setcounter{page}{1}
\maketitlesupplementary

\appendix
\section{Overview}
In this supplementary material, we provide the following relevant details that could not be included in the main paper:
\begin{enumerate}
    \item More details on LDM and ControlNet Implementation 
    \item Additional details of the Geometry Projection Module.
    \item Extended explanation of North-Aligned and Camera-Aligned setting of Ablation Study.
    \item Extended Ablation Results
    \item Additional Quantitative and Qualitative Results.
    \item Visualization of Failure Cases

\end{enumerate}

\section{Additional Details of LDM and ControlNet implementation on Cross-view Diffusion}

\subsection{Diffusion Models} 
\paragraph{Preliminary.}
Diffusion models \cite{sohldickstein2015deep, ho2020denoising, song2020denoising} are a class of latent variable models that have been proven to be superior to GANs in both unconditional and conditional image synthesis tasks \cite{dhariwal2021diffusion}. It is capable of learning a data distribution from an isotropic Gaussian distribution by reversing a diffusion process. 

Consider a forward diffusion process fixed to a Markov Chain that gradually adds Gaussian noise for a large number of timesteps T. The noising operator at each timestep $t \in \{1,\dots,T\}$ is defined as
\begin{equation}
    q(\mathbf{x}_t|\mathbf{x}_{t-1}) := \mathcal{N}(\mathbf{x}_t ;\sqrt{1-\beta_t}\mathbf{x}_{t-1}, \beta_t\mathbf{I}).
\end{equation}
By which we can compute the approximate posterior $q(\mathbf{x}_{1:T}|\mathbf{x}_0) := \prod^T_{t=1}q(\mathbf{x}_t|\mathbf{x}_{t-1})$ from $\mathbf{x}_0$ in the interested data distribution according to a variance schedule $\beta_1,\dots,\beta_T$ \cite{ho2020denoising}.

The reverse process is defined as a Markov Chain that performs sampling from $\mathbf{x}_T$ to $\mathbf{x}_0$. With each denoising step being expressed as a learned Gaussian transition parametrized by $\theta$ to approximate intractable true denoising distribution $q(\mathbf{x}_{t-1}|\mathbf{x}_t)$:
\begin{equation}
    p_\theta(\mathbf{x}_{t-1}|\mathbf{x}_t) := \mathcal{N}(\mathbf{x}_{t-1};\mu_\theta(\mathbf{x}_t,t), \Sigma_\theta(\mathbf{x}_t,t)).
\end{equation}

Ho et al. \cite{ho2020denoising} observe that the mean $\mu_\theta(\textbf{x}_t, t)$ of the denoising model can be represented by a noise estimator network $\epsilon_\theta(\mathbf{x}_t, t)$ to predict $\epsilon$ from $\mathbf{x}_t$, then sample $\mathbf{x}_{t-1}$:
    \begin{equation}
        \mathbf{x}_{t-1} =  \frac{1}{\sqrt{1-\beta_t}}\bigg(\mathbf{x}_t - \frac{\beta_t}{\sqrt{1-\Bar{\alpha}_t}}\epsilon_\theta(\mathbf{x}_t, t)\bigg) + \sigma_t\mathbf{z},
    \end{equation}
where $\mathbf{z} \sim \mathcal{N}(\mathbf{0},\mathbf{1})$ and $\Bar{\alpha} = \prod^t_{s=1}(1-\beta_s)$. 

Training of the denoiser network $\epsilon_\theta$
is performed with denoising score matching over multiple noise scales indexed by t \cite{song2020generative}:
\begin{equation}
    \mathcal{L}_{DM} := \mathbb{E}_{\mathbf{x},{\epsilon}\sim\mathcal{N}(\mathbf{0},\mathbf{I}), t}
    \bigg[\lambda_t\|\bm{\epsilon}-{\epsilon}_\theta(\mathbf{x}_t,t)\|^2_2\bigg]
\end{equation}

where $\mathbf{x}_t = \sqrt{\Bar{\alpha}_t}\mathbf{x}_0+ \sqrt{1-\Bar{\alpha}_t}\bm{\epsilon}$ and $\lambda_t = \frac{\beta^2_t}{2\sigma^2_t(1-\beta_t)(1-\Bar{\alpha}_t)}$, 
practically setting $\lambda_t = 1$ for improved sample quality \cite{ho2020denoising}.

\subsection{Difference between ControlNet and LDM models}
\label{sec:diff}
The original intention of implementing the ControlNet model with the pretrained SD model is to utilize its strong visual prior learned from millions of images. However, Stable Diffusion is essentially a text-to-image model. When there is no appropriate language prompt, our condition inputs serve as constraints during the image generation process, which might limit its generative capability. 

The main reason for us to report results from both ControlNet and LDM models is that we observe different performance that cannot be reflected solely on the quantitative metrics. As reported in \cref{fig:ldmvcn}, the LDM model trained from scratch can generate images with much diverse variation than the ControlNet model, in illumination, scene objects such as trees, boulders and buildings. Although the LDM model under-performs the ControlNet model in terms of quantitative metrics, its performances and synthesized images aligns better with our motivation to generate diverse image samples with the same condition. 

\paragraph{LDM Implementation.}
Incorporating our proposed Geometry-Guided Cross-View Condition, our conditional denoising step can be expressed as:
\begin{equation}
    p_\theta(\mathbf{z}_{t-1}|\mathbf{z}_t,c_{GCC}) := \mathcal{N}(\mathbf{z}_{t-1};\mu_\theta(\mathbf{z}_t,t, c_{GCC}), \Sigma_\theta(\mathbf{z}_t,t,c_{GCC})).
\end{equation}

Due to the computation resource limitation, our implementation deviates from configuration of the original Stable Diffusion model. We maintain the four blocks architecture of the LDM U-Net, but changed each block out channel size to $[240, 480, 960, 960]$, and also decreased the cross attention feature dimension from 1024 to 768.

\paragraph{ControlNet Implementation.}
As mentioned in Section \textcolor{red}{4} in the main paper, we have implemented a ControlNet \cite{zhang2023adding} version of our Cross-view diffusion pipeline for the effectiveness of our proposed Geometry-guided Cross-view Condition. Varying from the visual token sequence in the LDM \cite{dhariwal2021diffusion} version, we pixel-wisely align our condition with the encoded image latent and input it to the ControlNet module by reshaping the input tensor (see \cref{fig:overview}). The ControlNet module is a trainable copy of the encoder section of the LDM UNet, connected to the decoder section by zero convolution layers, whereas the LDM parameters are frozen. 

The pipeline is built upon pretrained Stable Diffusion 2.1 model~\cite{rombach2022highresolution}, where the prompt input to the LDM Model should be text embedding. During the training of the ControlNet Module, we set the text prompt to be an empty string to assure our generation results are unaffected by the text conditioning. In the future, we might explore the effect of combining both ControlNet and text conditions.

\section{Additional details of the Geometry Projection Module}
\label{sec:add_geo}
\paragraph{Geometric Projection Derivation for Ground Camera with Pin-hole Model}

%Similar to the ground-to-satellite localization tasks, 
In this paper, we consider the 3-DoF (Degree of Freedom) ground camera pose for the KITTI \cite{geiger2013vision} dataset, \ie, 
% In cross-view localization tasks, we only care about 3 DoFs of the ground camera pose, 
the 1-DoF azimuth angle $\phi \in [-\pi, \pi]$ and 2-DoF translation along the latitude and longitude directions. Let $\mathbf{R} = \big(\begin{smallmatrix}
  \cos\phi & 0 & -\sin{\phi}\\
  0 & 1 & 0 \\
  \sin{\phi} & 0 & \cos\phi
\end{smallmatrix}\big)$ and $\mathbf{t} = [t_x, 0, t_z]^T$ be the relative rotation and translation from real ground camera coordinate system to the world coordinate system and $\mathbf{K}$ be the ground camera intrinsics. 

The back-projection from a pixel on a pin-hole camera image plane to the world coordinate system can be expressed as

\begin{equation} \label{eq:backproj}
    [x,y,z]^T = w \mathbf{R} \mathbf{K}^{-1}[u_g , v_g , 1]^T +  \mathbf{R}\mathbf{t}
\end{equation}
where $w$ is a scare factor.\\

By combining \cref{eq:ortho} from \cref{sec:proj} and \cref{eq:backproj} above, we can derive the mapping from a ground-view pixel $(u_g,v_g)$ to a satellite pixel $(u_s,v_s)$ as

\begin{equation}
    \begin{bmatrix}
        u_s \\
        v_s \\
        z
    \end{bmatrix} = 
    \begin{bmatrix}
        \frac{1}{\gamma} & 0 & 0 \\
        0 & \frac{1}{\gamma} & 0 \\
        0 & 0 & 1
    \end{bmatrix}
     \begin{pmatrix}
         w \mathbf{R} \mathbf{K}^{-1}  
    \begin{bmatrix}
        u_g \\
        v_g \\
        1
    \end{bmatrix} +\mathbf{R}\mathbf{t}
     \end{pmatrix} + \begin{bmatrix}
        u_s^0 \\
        v_s^0 \\
        0
    \end{bmatrix}.
\end{equation}
The above projection is defined on ground plane homography, $w$ is therefore computed based on the assumption of fixed camera height $y_c$.
Similarly, we can derive the mapping from an satellite pixel to a ground image pixel

\begin{equation}
\scalebox{0.9}{
$\begin{bmatrix}
        u_g \\
        v_g
    \end{bmatrix} = 
    \begin{bmatrix}
        f_x\frac{[(v_s-v^0_s)+t_x]\cos{(-\phi)} - [(u_s-u^0_s)+t_z]\sin{(-\phi)}}{[(v_s-v_s^0)+t_x]\sin{(-\phi)}+[(u_s-u^0_s)+t_z]\cos{(-\phi)}}\\
        f_y\frac{h}{\gamma\big[[(v_s-v^0_s)+t_x]\sin{(-\phi)}+[(u_s-u^0)+t_z]\cos{(-\phi)}\big]}  
    \end{bmatrix} + \begin{bmatrix}
        u^0_g \\ v^0_g
    \end{bmatrix},$
}
\end{equation}

where $f_x$ and $f_y$ denote the ground camera focal length along $u$ and $v$ directions, respectively, $h$ is the height of pixel $(u_s, v_s)$ above the ground plane.

\section{Extended Explanation of North-Aligned and Camera-Aligned setting}

\label{sec:align}

\begin{figure}[h]
   \centering
   \includegraphics[width=0.8\linewidth]{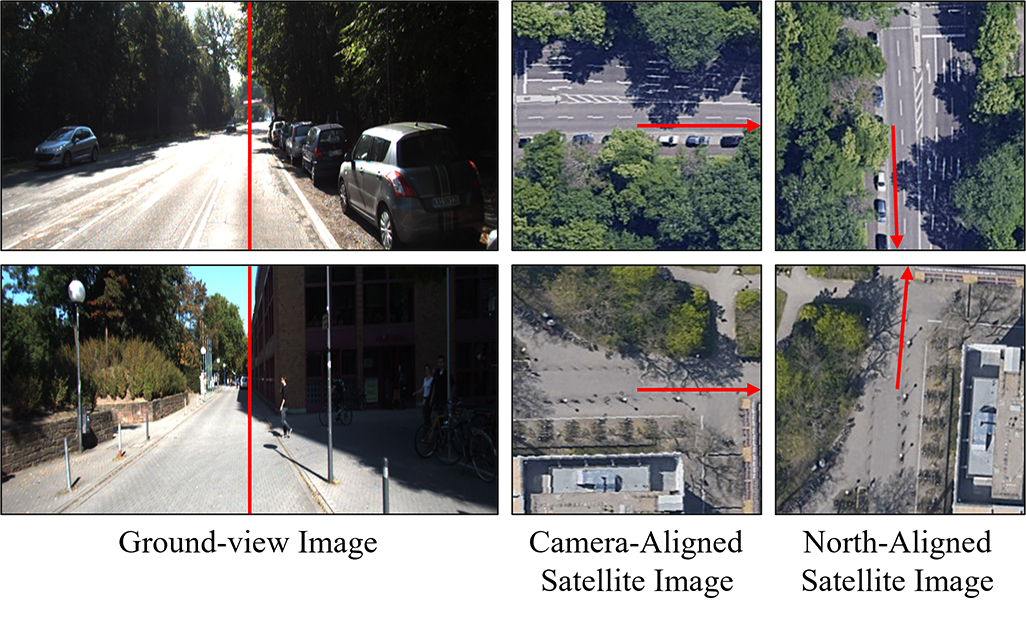}
   \caption{Example of Camera-Aligned and North-Aligned samples, the red arrows in the satellite views indicate the orientation of the ground camera.}
   \label{fig:align}
\end{figure}

\begin{figure*}[h]
  \centering
  \includegraphics[width=0.9\linewidth]{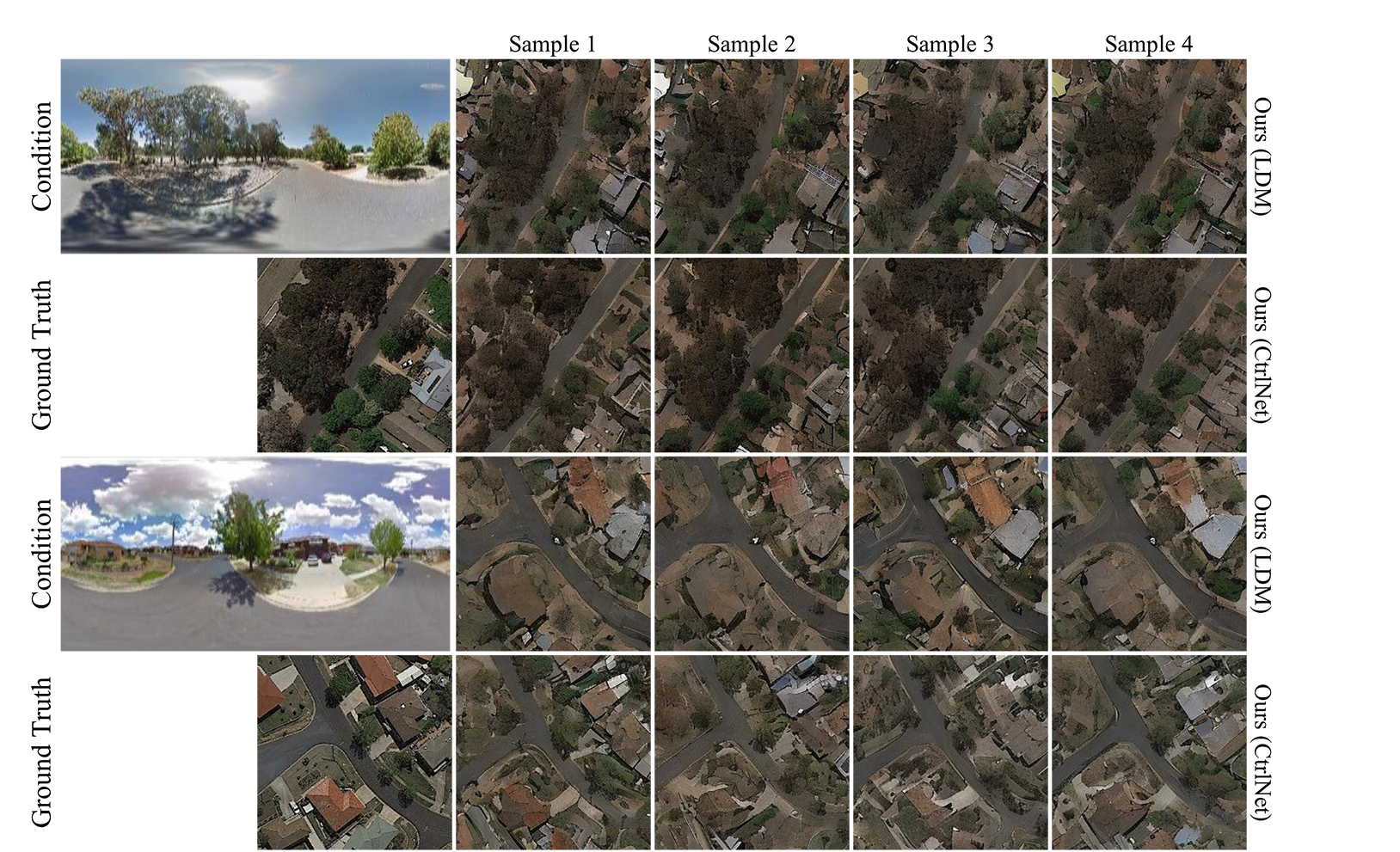}
  \caption{Ablation for sample generated by our LDM model and our ControlNet model given the same condition, on \textbf{Grd2Sat} task.}
  \vspace{-0.3cm}
  \label{fig:ldmvcn2}
\end{figure*}

As mentioned in 
\cref{sec:ablation} in the main paper, we presented ablation study results for camera-aligned and north-aligned setup on the KITTI dataset. As illustrated in \cref{fig:align}, %we illustrate the orientation relationship between the satellite images and the ground-view images. 
under the Camera-Aligned setting, the orientation of the ground-view image is always aligned in the same direction on the satellite view. When the satellite images are North-aligned, the orientation relationship between the satellite images and the ground-view image changes between pairs, which yields pose ambiguity between the cross-view image pairs that hinders the models' learning ability as reported in the \cref{table:cond_ablat_kitti} of the main paper. However, our experiment show that the model with projected feature condition suffers less performance drop under the North-aligned setting comparing to the image condition, which can effectively mitigate the influence of pose ambiguity.

\section{Further Ablation results on the Generative Ability of Models}
\label{sec:fur_abla}
In \cref{fig:ldmvcn2}, we show the qualitative ablation on the \textbf{Grd2Sat} task with generated samples from both LDM and ControlNet Models. As stated in the main paper, the generative ability for the \textbf{Grd2Sat} is limited by the variability of the data itself, therefore, we do not see much diversity in the generated samples compared to the samples from the \textbf{Sat2Grd} task.

\section{Additional Qualitative and Quantitative Results}
\begin{figure*}[h]
   \centering
   \includegraphics[width=0.8\linewidth]{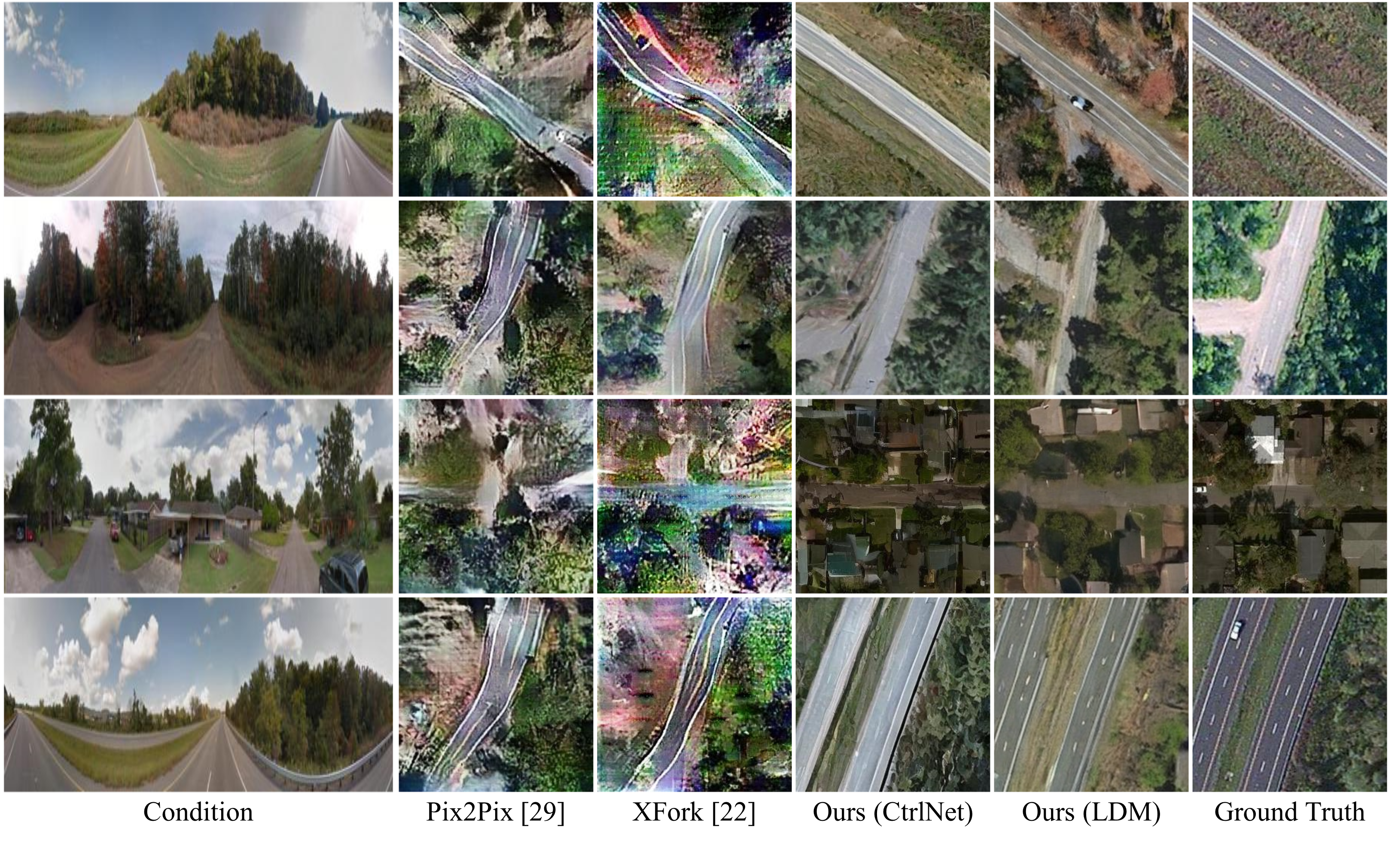}
   \caption{Qualitative comparisons of our results on the \textbf{Grd2Sat} task, on CVUSA dataset.}
   \label{fig:cvusa_g2s}
\end{figure*}

\begin{table}[h]
    \centering
    \caption{Overall Evaluation without sky region, on CVUSA, best in \textbf{bold}}
      \label{table:cvusa_eval}
      \begin{tabular}{c c c c c}
        \toprule
         Method & PSNR$\uparrow$ & SSIM$\uparrow$ & $P_{\text{alex}}\downarrow$ & $P_{\text{squeeze}}\downarrow$\\
         \midrule
         Sat2Density & 14.528 & 0.2389 & 0.3958 & 0.3084\\
         Ours(LDM) & 14.791 & \textbf{0.2908} & 0.3867 & \textbf{0.3074} \\
         Ours(CtrlNet)& \textbf{14.879} & 0.2725 & \textbf{0.3861} & 0.3090 \\
        \bottomrule
    \end{tabular}
\end{table}
In \cref{fig:cvusa_g2s}, we include qualitative comparisons of \textbf{Grd2Sat} results with existing methods on the CVUSA dataset. In \cref{table:cvusa_eval}, we conduct another evaluation with the sky regions excluded, evaluating only the shared region between the ground-view and satellite-view on the ground-level. Our results outperform Sat2Density in all metrics, showing that we are able to generate more geometrically and semantically aligned images with diversity.
\section{Visualization of Failure Cases}
\begin{figure*}[t]
   \centering
   \includegraphics[width=0.8\linewidth]{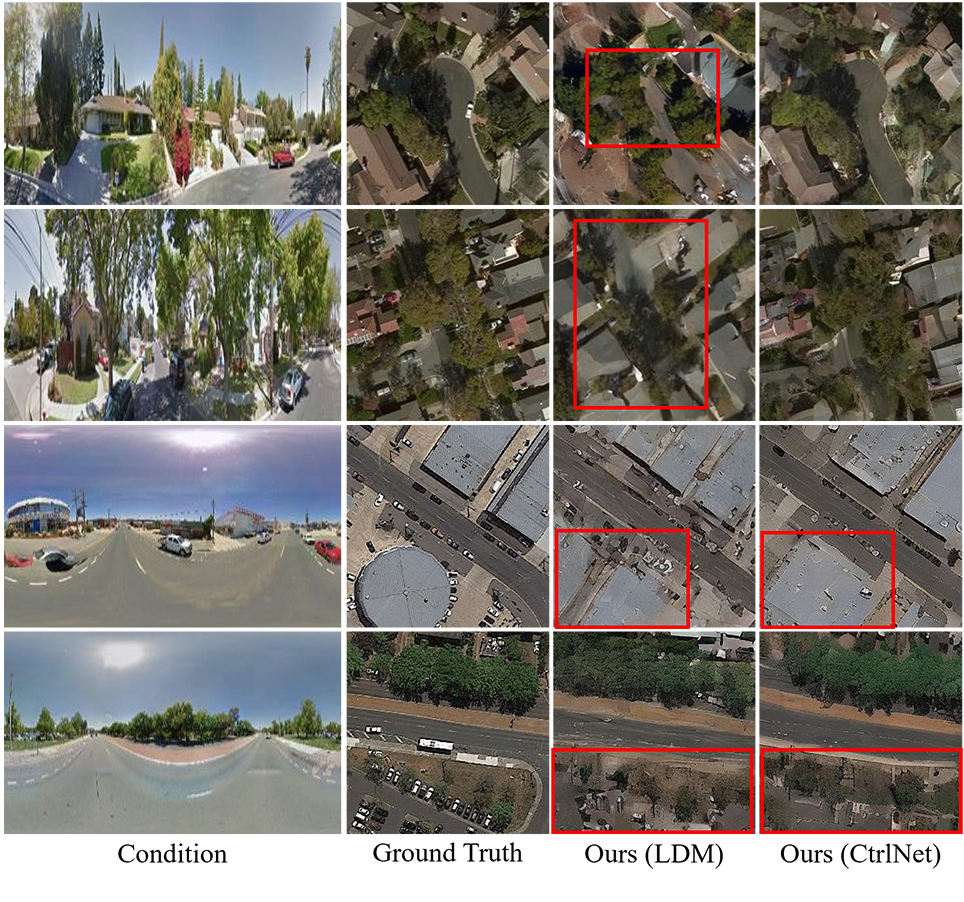}
   \caption{Some Failure cases on \textbf{Grd2Sat} task, on both CVUSA and CVACT datasets. We mainly visualize failure cases in \textbf{Grd2Sat}, as it is a much challenging task to learn and recover geometric and textural information by geometric projected feature alone, due to presence of limited range of sight (row 4), occlusion (row 2 and 4) and shape ambiguity (row 1 and 3). }
   \label{fig:fail}
   
\end{figure*}

In \cref{fig:fail}, we show some typical failure cases from \textbf{Grd2Sat}, on both CVUSA and CVACT datasets. The first two rows are samples from CVUSA, and last two rows are samples from CVACT.  

In the first two rows, samples generated by our LDM model failed to reconstruct the true street structure, this might due to the model failed to pick up structural information from the given condition. As summarized in the main paper, our ControlNet version generally outperforms our LDM version in the \textbf{Grd2Sat} task, this might due to the stronger supervision from features that are pixel-aligned with the image latent. The samples generated in the third row failed to recover the shape of the round building, where the building shape can not be recognized simply by projecting the ground-view panorama. In the fourth row, the samples failed to generate the correct road structure at end of the road and also the car park behind the pedestrian walkway due to limited range of sight and occlusion in the ground-view.

